\documentclass[10pt,twocolumn,letterpaper]{article}

\usepackage[pagenumbers]{cvpr} 

\graphicspath{{./}}
\usepackage[dvipsnames]{xcolor}

\definecolor{cvprblue}{rgb}{0.21,0.49,0.74}
\usepackage[pagebackref,breaklinks,colorlinks]{hyperref}
\usepackage{multirow}
\usepackage{diagbox}
\usepackage{soul}
\usepackage{color, xcolor}

\def\paperID{15176} 
\def\confName{CVPR}
\def\confYear{2024}

\title{PR-NeuS: A Prior-based Residual Learning Paradigm for Fast Multi-view Neural Surface Reconstruction}

\def\authorBlock{
    \textbf{Jianyao Xu}$^1$\footnotemark[1] \qquad
    \textbf{Qingshan Xu}$^2$\footnotemark[1] \qquad
    \textbf{Xinyao Liao}$^1$\qquad
    \textbf{Wanjuan Su}$^1$\\
    \textbf{Chen Zhang}$^1$\qquad
    \textbf{Yew-Soon Ong}$^{2,3}$\qquad
    \textbf{Wenbing Tao}$^1$\footnotemark[2]\\  
    $^1$Huazhong University of Science and Technology\\
    $^2$Nanyang Technological University \qquad
    $^3$A*STAR, Singapore
}
\author{\authorBlock}

\begin{document}

\twocolumn[{
\renewcommand\twocolumn[1][]{#1}
\maketitle
\centering
\vspace{-0.5cm}
 \includegraphics[width=\linewidth]{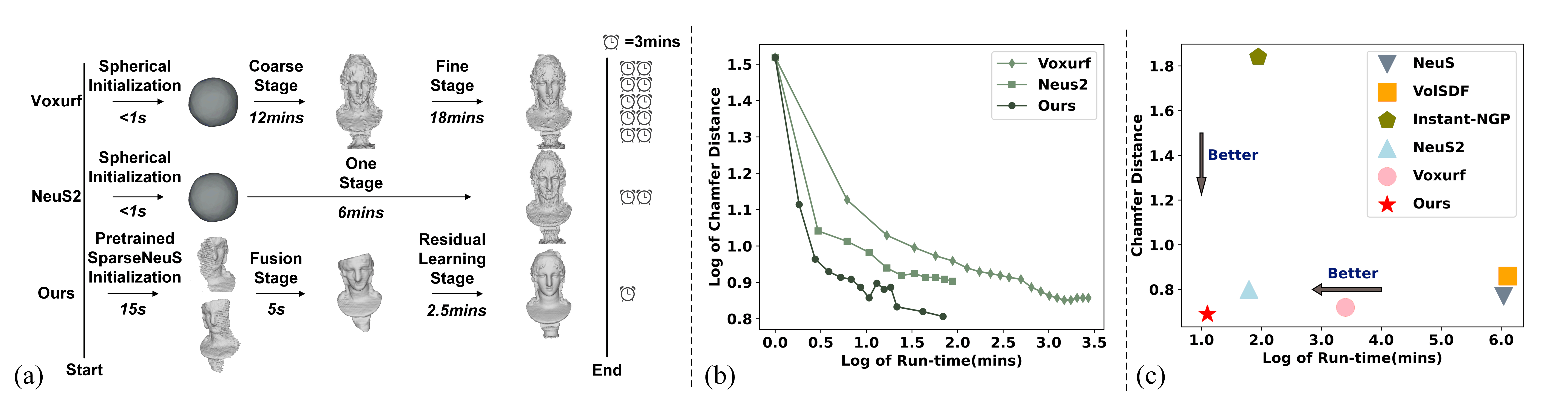}
 
 \captionsetup{type=figure}
\caption{We show comparisons about \textit{(a) core idea}, \textit{(b) convergence speed}, \textit{(c) quality and efficiency} among different methods.
}
\vspace{0.2cm}
\label{fig:teaser_demo}
}]

\renewcommand{\thefootnote}{\fnsymbol{footnote}}
\footnotetext[1]{Equal contribution}
\footnotetext[2]{Corresponding author}
\renewcommand{\thefootnote}{\arabic{footnote}}

\maketitle
\begin{abstract}
Neural surfaces learning has shown impressive performance in multi-view surface reconstruction. However, most existing methods use large multilayer perceptrons (MLPs) to train their models from scratch, resulting in hours of training for a single scene. Recently, how to accelerate the neural surfaces learning has received a lot of attention and remains an open problem. In this work, we propose a prior-based residual learning paradigm for fast multi-view neural surface reconstruction. This paradigm consists of two optimization stages. In the first stage, we propose to leverage generalization models to generate a basis signed distance function (SDF) field. This initial field can be quickly obtained by fusing multiple local SDF fields produced by generalization models. This provides a coarse global geometry prior. Based on this prior, in the second stage, a fast residual learning strategy based on hash-encoding networks is proposed to encode an offset SDF field for the basis SDF field. Moreover, we introduce a prior-guided sampling scheme to help the residual learning stage converge better, and thus recover finer structures. With our designed paradigm, experimental results show that our method only takes about 3 minutes to reconstruct the surface of a single scene, while achieving competitive surface quality. Our code will be released upon publication.
\end{abstract}
    
\section{Introduction}
\label{sec:intro}

Reconstructing 3D scene surfaces from multiple 2D images is of great significance in many real-world applications, such as AR/VR, autonomous driving and so on. Most of traditional methods use multi-view stereo algorithms \cite{colmap,xu2019multi,xu2022multi, barnes2009patchmatch,furukawa2009accurate,galliani2016gipuma} to produce dense point clouds and then convert the point clouds into a surface model via surface reconstruction algorithms \cite{Kazhdan2013Screened,kazhdan2020poisson,huang2019variational,calakli2011ssd,fuhrmann2014floating}. However, their reconstruction quality is limited due to the accumulated errors caused by the long processing pipeline. With the rise of
neural volume rendering \cite{max1995optical,nerf,reiser2023merf,johari2022geonerf,niemeyer2022regnerf,wang2021ibrnet,zhang2020nerf++,tancik2022block}, implicit neural surfaces learning has been applied to multi-view surface reconstruction. Represented by NeuS \cite{neus} and VolSDF \cite{volsdf}, geometric surfaces are modeled as the signed distance function (SDF) \cite{vf:nerf:deepsdf} field encoded by a neural network. By transforming the SDF into density representations, differential volume rendering is used to render image colors. 
In this way, the photometric loss is minimized to directly learn the final geometry without intermediate transformations of the geometry representation. 
As a result, neural surfaces learning has shown impressive performance on surface reconstruction tasks.

In general, most neural surfaces learning methods use \emph{large multilayer perceptrons (MLPs)} to train their models \emph{from scratch}, resulting in a lengthy training time for a single scene. This severely impedes the usability of these existing methods in practice. To speed up the training time, some methods \cite{sun2022direct,fridovich2022plenoxels,ngp,voxurf} exploit an efficient grid representation with a shallow MLP to model implicit neural fields. Thanks to the direct grid optimization, these methods greatly reduce training time (about 30 minutes). On the other hand, inspired by generalizable neural radiance fields (NeRF) \cite{yu2021pixelnerf,chen2021mvsnerf,Schwarz20neurips_graf,Trevithick_2021_ICCV}, some methods \cite{sparseneus,volrecon} have been proposed to achieve fast generalizable reconstruction by leveraging image features. These methods specify a reference viewpoint to construct cost volumes from most related neighboring image features, thus obtaining scene priors to quickly learn implicit fields. However, their reconstructed geometries usually fail to describe the entire scene due to their local scene perception capability. Moreover, their reconstruction quality is limited by the cost volume resolution.

In this work, we leverage generalization priors to achieve \emph{fast} and \emph{complete} geometry reconstruction. To this end, we propose a prior-based residual learning paradigm for fast multi-view neural surface reconstruction, called \emph{PR-NeuS}. This paradigm consists of two fast optimization stages (Fig.~\ref{fig:teaser_demo}(a)). In the first stage, we propose to leverage generalization models to generate a global geometry prior. Since the generalization models can only output local incomplete geometry fields, we first compute multiple local SDF fields from different sampling viewpoints. To integrate these incomplete geometries into a complete one, an efficient local SDF fields fusion strategy is introduced to obtain a basis SDF field. 
In the second stage, we propose a fast residual learning strategy to predict an offset SDF field for the basis SDF field. This strategy leverages hash-encoding grid representations to guarantee efficiency. In addition, we design a prior-guided sampling scheme to help the residual learning stage converge better. This scheme not only guides the optimization to pay more attention to the spatial points around the coarse surfaces, but also introduces randomness to escape local optima. As a result, our residual learning strategy can recover finer structures. 

Our paradigm takes the best of the two worlds, generalization models and grid-based representations. Compared with existing generalizable methods, our introduced fusion strategy allows to generate global geometry priors and the residual learning network is able to quickly recover details based on the priors. Compared with existing grid-based methods, our method leverages geometry priors to only learn offset SDF fields, thus not needing to train models from scratch for each scene. Our method can, therefore, achieve very fast surface reconstruction, taking only about 3 minutes for a single scene reconstruction. Moreover, our reconstructed surface quality is competitive with existing neural surfaces learning methods with large MLPs.

In summary, our work has the following contributions:
\begin{itemize}
    \item A prior-based residual learning paradigm is proposed for fast neural surfaces learning from multi-view images. This paradigm enables our method to take only about 3 minutes to reconstruct a scene while achieving competitive surface quality (Fig.~\ref{fig:teaser_demo}).
    \item A multiple local SDF fields fusion strategy is introduced to generate coarse global geometries. This strategy can efficiently provide more complete geometry priors to guide the subsequent residual learning stage. 
    \item A Fast SDF residual learning strategy is proposed to learn an offset SDF field for the global geometry priors. Moreover, an efficient prior-guided sampling scheme is presented to facilitate the recovery of fine structures. This helps our method reconstruct high-quality surfaces at a very low cost.
\end{itemize}

\section{Related Work}
\label{sec:related work}

\noindent\textbf{Neural surfaces learning from multi-view images} 
Neural radiance field is a brand-new way to encode the geometry and appearance of a 3D scene by neural networks. As the NeRF \cite{nerf} first applies neural radiance field in novel view synthesis tasks, a plethora of papers have searched on how to use neural radiance fields to reconstruct the surface of a scene \cite{spn:14:sdfdiff,spn:50,spn:31:dvr,spn:17,spn:22,neus,volsdf,unisurf,spn:52:mvsdf,spn:6:neuralwarp,geoneus}.
Unisurf \cite{unisurf} brings the occupancy value \cite{vf:nerf:occ_net} into the volume rendering to improve the quality of 3D geometry. 
NeuS \cite{neus} and VolSDF \cite{volsdf} build the bridge between SDF \cite{vf:nerf:deepsdf} and the volume rendering and achieve nice results. 
Geo-Neus \cite{geoneus} introduces two explicit multi-view constraints
and reconstructs high-quality surfaces.
Though these methods can recover good geometries, their heavy networks composed of large MLPs require hours for per-scene optimization.  
Based on these nice works, our work investigates how to achieve fast neural surfaces learning while guaranteeing competitive surface quality.  

\noindent\textbf{Fast geometry encoding.} 
There exist two kinds of methods to achieve fast geometry encoding: fast per-scene optimization with grid-based representations and generalization models by incorporating learned priors. To speed up the geometry encoding in NeRF, some methods \cite{fridovich2022plenoxels,sun2022direct} use grid-based representations to replace large MLPs. This direct grid optimization greatly reduces per-scene training time. 
Based on the grid-based representation, Instant-NGP \cite{ngp} further proposes the multi-resolution of hash-tables for encoding the geometry and appearance, whose time of per-scene training drops from several hours to several minutes compared with MLPs encoding. 
The Voxurf \cite{voxurf} introduces a voxel-based surface representation for fast multi-view 3D reconstruction, but still needs almost one hour for per-scene training on a consumer GPU. 
The NeuS2 \cite{neus2} achieves good reconstruction within less than 10 mins, by combining \cite{ngp} and \cite{neus} as well as applying lots of low-level code optimization using Cuda and C++. 
Another line of research combines learned cost volumes with neural rendering to construct generalization NeRF models \cite{yu2021pixelnerf,chen2021mvsnerf}. 
Inspired by this, SparseNeuS \cite{sparseneus} and VolRecon \cite{volrecon} are the generalizable surface reconstruction methods that can quickly obtain geometry priors from learned cost volumes. They don't need per-scene optimization, but their reconstructed surface quality becomes worse due to their limited local geometry priors acquisition. Our method combines the merits of fast generalizable geometric priors and fast hash-table encoding, so as to achieve not only high-quality surface reconstruction but also less than 5 minutes' optimization.

\begin{figure*}[t]
\centering
\includegraphics[width=1.0\linewidth]{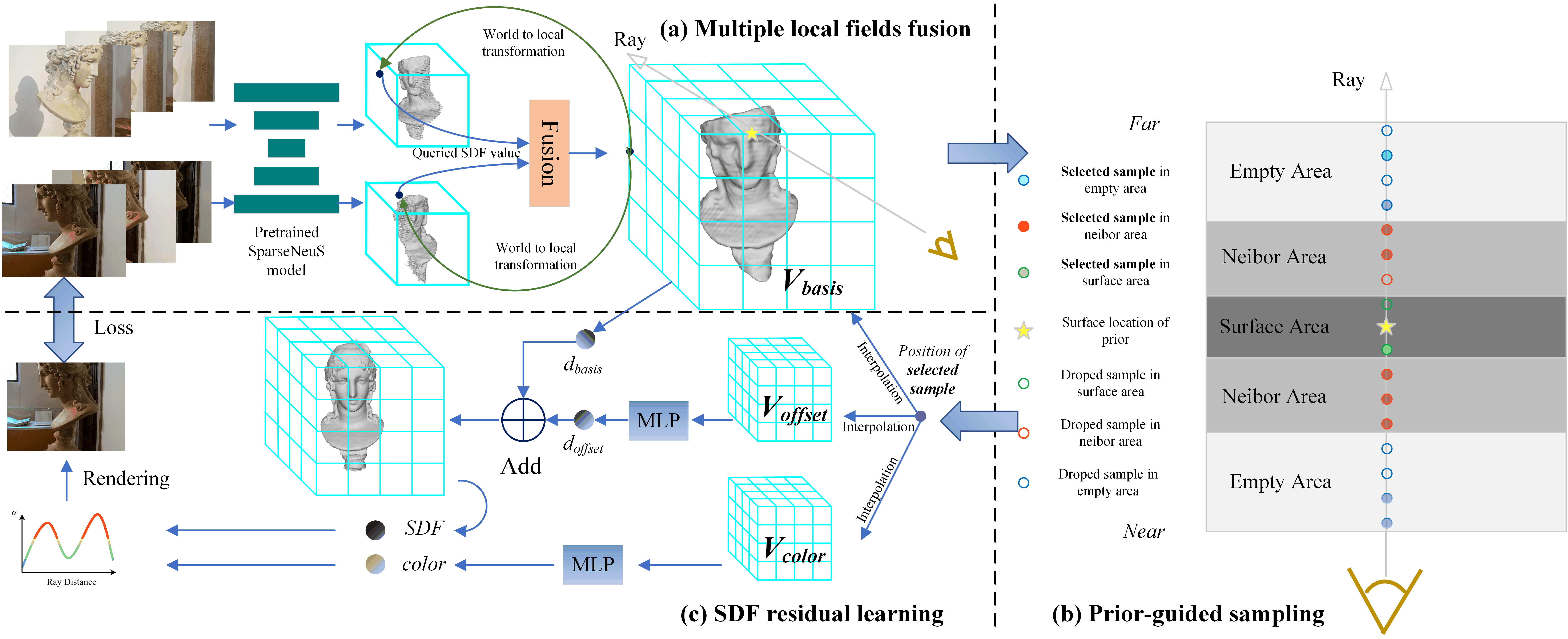}
\caption{\textbf{Overview.} First, we adopt the multiple local SDF fields fusion for priors obtained from pretrained SparseNeuS model, and get a global basis SDF field $V_\text{basis}$ as a result. Based on the global priors, we design the prior-guided sampling to assign more samples near the surface. In SDF residual learning, we use an offset network consists of hash-table and shallow MLP to predict the offset SDF values $d_{\text{offset}}$, which will be added to the basis SDF value $d_\text{basis}$ interpolated from $V_\text{basis}$ to get the final SDF value for subsequent volume rendering.}
\label{fig:pipeline}  
\end{figure*}

\section{Method}

Fig.~\ref{fig:pipeline} shows the flowchart of PR-NeuS.  
This is a prior-based residual learning paradigm. 
First, the multiple local SDF fields fusion is used to obtain a basis SDF field as geometric priors. Then, the prior-guided sampling is applied to sample points. Finally, by combining our sampling strategy, the SDF residual learning is adopted to optimize an offset SDF field for the basis SDF field via volume rendering.

\subsection{Multiple Local SDF Fields Fusion}

In our prior-based residual learning paradigm, we hope to obtain global geometric priors quickly to guide the subsequent residual learning stage. To this aim, we introduce a multiple local SDF fields fusion strategy to leverage generalization models to quickly obtain global geometry priors. 

Given $N$ input images with their known camera parameters, we select $M$ groups of sparse inputs from $M$ different viewpoints. We use SparseNeuS, a generalization method, to compute $M$ rough SDF fields. These rough SDF fields are in
their respective local camera coordinate systems and may contain some outliers. To fuse them into a unified SDF field, we define a voxel grid for a scene in a unified world coordinate system. Each vertex of this voxel grid stores an SDF value. Then, the SDF value of any point in 3D space can be obtained by trilinear interpolation of its eight neighboring vertices, thereby constructing a global SDF field. 

More specifically, we regard the vertices of the voxel grid as a point cloud $\boldsymbol{P}_\text{vert}$ , then map the point cloud $\boldsymbol{P}_\text{vert}$ to $M$ SDF fields through coordinate transformation. In this way, a point $\boldsymbol{p}$ in the point cloud $\boldsymbol{P}_\text{vert}$ will query $M$ SDF values from $M$ local SDF fields. 
Its SDF value with respect to the the \textit{i-th} local SDF field $V_\text{local}^i$ is computed as:
\begin{equation}\label{eq：query} 
d^{i}(\boldsymbol{p}) = \text{interp}\left( \boldsymbol{S}_i \cdot \boldsymbol{R}_i \cdot \boldsymbol{p}, V_\text{local}^i \right), i \in \{1,2 \ldots M \}, 
\end{equation}
where $\text{interp}(\cdot, \cdot)$ denotes trilinear interpolation, $\boldsymbol{R}_i$ is the transformation matrix from the world coordinate system to the \textit{i-th} local coordinate system and $\boldsymbol{S}_i$ is the scaling matrix from the \textit{i-th} local coordinate system to the normalized coordinate system. 
To fuse the SDF values of $M$ local SDF fields, our fusion strategy is based on the principle that preserving the SDF value of the vertex near the surface as much as possible. 
Our fusion function behaves as follows:
\begin{equation}\label{eq:fusion}
\hat{d}(\boldsymbol{p}) = d^{I}(\boldsymbol{p}), I = \text{argmin} \left(|d^{1}(\boldsymbol{p})|,|d^{2}(\boldsymbol{p})|,\ldots,|d^{M}(\boldsymbol{p})| \right).
\end{equation}
Then, we can get a raw global SDF field grid $V_\text{global} = \{\hat{d}(\boldsymbol{p})\ |\ \boldsymbol{p} \in \boldsymbol{P}_\text{vert}\}$. 
Since there inevitably retain many noises in the local SDF fields, the inferred global SDF field is far from smooth and continuous. 
Therefore, we apply the Gaussian convolution to smooth the global SDF field $V_\text{global}$. The final obtained global SDF field is defined as our basis SDF field $V_\text{basis}$. This provides clean and complete geometric priors for the subsequent SDF residual learning process.

\subsection{Prior-guided Sampling}

After obtaining the basis SDF field, our SDF residual learning strategy will leverage volume rendering to compute an offset SDF field. In the vanilla volume rendering process, a stratified sampling strategy is used to sample spatial points on a ray \cite{nerf}. This sampling introduces many unnecessary spatial point queries, hindering the efficiency of volume rendering. Based on our obtained basis SDF field, we propose an efficient prior-guided sampling strategy to facilitate our subsequent SDF residual learning. 

Our goal here is to sample more spatial points around the coarse surface as much as possible, while introducing randomness to avoid falling into local optima. The basis SDF field $V_\text{basis}$ explicitly indicates the area near the surface. 
Therefore, we use an occupancy grid derived from $V_\text{basis}$ to guide the sampling. Specifically, we first use the marching cube algorithm \cite{lorensen1998marching} to extract a mesh from $V_\text{basis}$. Then we divide the entire scene space into three parts: 1) $A_{2}$: The area occupied by the initial rough surface mesh. 2) $A_{1}$: The neighbor area near that occupied by the initial rough surface mesh. 3) $A_{3}$: The other empty area excluding $A_{1}$ and $A_{2}$. Furthermore, we define the priority order of these sampling areas as $A_{1}>A_{2}>A_{3}$. Since there is a certain small error between the rough surface and the real surface, the real surface is most likely in neighbor area $A_{1}$. So we shall sample more points on $A_{1}$ to reserve sufficient space for the subsequent optimization of the rough surface. Therefore the priority order is $A_{1}>A_{2},A_{3}$. As for the rough surface area $A_{2}$, its distance from the real surface is relatively smaller compared to the empty area $A_{3}$, thus the priority order between $A_{2}$ and $A_{3}$ is $A_{2}>A_{3}$.

To reflect the sampling priority in the distribution of sampling points, we first conduct equidistant sampling on a ray. Each sampling point $s_i$ must belong to one of these three areas, and its area tag is $t_i \in \{1, 2, 3\}$. 
Then we apply the random dropout strategy to control the sampling distribution of three areas. Random dropout means that for any equidistant sampling point $s_i$, it will be abandoned with a certain probability $(1-\boldsymbol{P}(s_i))$. $\boldsymbol{P}(s_i)$ is defined as:
\begin{align}\label{eq:reserve-probability}
  \boldsymbol{P}(s_i) = \text{min}(1.0,\beta_{t_i} \cdot \frac{N_\text{ref}}{N(A_{t_i})}),
\end{align}
where $\beta_{t_i}$ denotes the scalar weight of priority $t_{i}$, with $\beta_1 > \beta_2 > \beta_3$, $N(A_{t_i})$ denotes the number of voxels belonging to the area $A_{t_i}$. The $N_\text{ref}$ denotes the reference number of voxels, and we set $N_\text{ref}=N(A_2)$. The quantity distribution of three areas $A_1, A_2, A_3$ varies greatly in different scenes. To ensure that the dropout probability can be adapted to different scenes, the reserve-probability $\boldsymbol{P}(s_i)$ of sampled point $s_i$ is set to be inversely related to the quantity of area $A_{t_i}$. The larger $N(A_{t_i})$, the smaller $\boldsymbol{P}(s_i)$. In this way, the final sampled point set $S$ is defined as:
\begin{equation}\label{eq:sample}
    S = \{s_i | m(s_i) = 1\}, 
\end{equation}
\begin{equation}\label{eq:mask}
    m(s_i) = \left\{
\begin{aligned}
  1 \qquad rand < \boldsymbol{P}(s_i) \\
  0 \qquad rand > \boldsymbol{P}(s_i) \\
\end{aligned},
\right. 
\end{equation}
where $rand$ represents a random value generated from the uniform distribution $U(0,1)$. Thanks to Eq.~\eqref{eq:reserve-probability}-\eqref{eq:mask}, the quantity of sampling in area $A_1, A_2, A_3$ can keep the ratio of $\beta_1:\beta_2:\beta_3$ all the time.

\subsection{SDF Residual Learning}

In this stage, we aim to recover fine geometries quickly based on the rough basis SDF field. Thus, we propose an SDF residual learning strategy. We take advantage of a hash-encoding network to predict an offset SDF field for the basis SDF field by volume rendering. In this way, we can make full use of the power of both the previously obtained geometric priors and efficient hash-encoding. 

\noindent\textbf{Offset SDF network.} Given the basis SDF field $V_\text{basis}$, for any 3D position $\boldsymbol{p}$, we apply a hash-encoding network $\{H_{\Omega}, MLP_\Theta\}$ to predict its offset SDF value $d_\text{offset}(\boldsymbol{p})$ with respect to its basis SDF value $d_\text{basis}(\boldsymbol{p})$. The $d_\text{basis}(\boldsymbol{p})$ is directly interpolated from $V_\text{basis}$. Specifically, we map $\boldsymbol{p}$ to multi-resolution hash-encodings $H_{\Omega}(\boldsymbol{p})$ with learnable hash-table entries $\Omega$. 
As $H_{\Omega}(\boldsymbol{p})$ is a multi-resolution informative encoding of 3D spatial position, it contains more geometry information from coarse level to fine level compared with fixed-resolution grid encoding. 
The MLPs used for decoding can be very shallow, which results in more efficient rendering and training without compromising quality.
\begin{align}\label{eq:offset}
d_\text{offset}(\boldsymbol{p}) = MLP_{\Theta}(\boldsymbol{p}, H_{\Omega}(\boldsymbol{p})), 
\end{align}
where $\Omega, \Theta$ of $H_{\Omega}, MLP_{\Theta}$ are trainable parameters. 
The final SDF value $d(\boldsymbol{p})$ is composed of $d_\text{basis}(\boldsymbol{p})$ and $d_\text{offset}(\boldsymbol{p})$:

\begin{align}
d(\boldsymbol{p})= d_\text{basis}(\boldsymbol{p})+d_\text{offset}(\boldsymbol{p}).
\end{align}
Since $d_\text{basis}(\boldsymbol{p})$ provides a relatively good prior for the SDF residual learning, we assign the offset SDF value at any position to 0 at the beginning of learning, that is $d_\text{offset}(\boldsymbol{p})=0$.

\noindent\textbf{Normal vector.} The normal vector $n=\nabla d(\boldsymbol{p})$ plays an crucial role in volume rendering. 
In our residual learning stage, the SDF value $d(\boldsymbol{p})$ consists of two parts, resulting in that the normal vector $n(\boldsymbol{p})$ is computed as: $n(\boldsymbol{p})=\nabla(d_\text{basis}(\boldsymbol{p})+d_\text{offset}(\boldsymbol{p}))=\nabla d_\text{basis}(\boldsymbol{p}) + \nabla d_\text{offset}(\boldsymbol{p})=n_\text{basis}(\boldsymbol{p})+n_\text{offset}(\boldsymbol{p})$,
where $n_\text{basis}(\boldsymbol{p})$ and $n_\text{offset}(\boldsymbol{p})$ are the gradients of $d_\text{basis}(\boldsymbol{p})$ and $d_\text{offset}(\boldsymbol{p})$ respectively. Since $d_\text{basis}(\boldsymbol{p})$ is fixed, $n_\text{basis}(\boldsymbol{p})$ is a constant and doesn't participate in training.  

\noindent\textbf{Color network.} We use another hash-encoding network $\{ H_{\Omega_2}, MLP_{\Theta_2} \}$ to encode the color field. For any 3D point $\boldsymbol{p}$ with embedded viewing direction $\boldsymbol{v}$, the color network outputs RGB color values as: 
\begin{align}\label{eq:color network}
c(\boldsymbol{p}) = MLP_{\Theta_2}(\boldsymbol{p},H_{\Omega_2}(\boldsymbol{p}), \boldsymbol{v},d(\boldsymbol{p}),n(\boldsymbol{p})). 
\end{align}

\noindent\textbf{Volume rendering.} Following the NeuS \cite{neus}, the color of ray $\boldsymbol{r}$ is rendered by accumulating the colors of $K$ samples along the ray as: $C(\boldsymbol{r})=\sum_{i=1}^K T_i \alpha_i c_i$,  
where $\alpha_i=\text{max}(-\frac{\Phi_s^{'}(d_i)}{\Phi_s(d_i)}, 0)$ is opacity derived from SDF values, $T_i=\prod_{j=1}^{i-1}(1-\alpha_i)$ is accumulated transmittance and $\Phi_s$ is the cumulative distribution of logistic distribution. Similarly, we render the normal of the ray $\boldsymbol{r}$ as: $N(\boldsymbol{r})=\sum_{i=1}^K T_i \alpha_i n_i$.

\subsection{Optimization}

In the SDF residual learning, the final loss is defined as:
\begin{align}\label{eq:total}
  \mathcal{L}_\text{total} = \mathcal{L}_\text{rgb} + \lambda_1 \mathcal{L}_\text{patch} + \lambda_2 \mathcal{L}_\text{eik} + \lambda_3 \mathcal{L}_\text{normal},
\end{align}
where $\lambda_1$, $\lambda_2$ and $\lambda_3$ are weights to control the importance of different terms.

$\mathcal{L}_\text{rgb}$ is the reconstruction loss by comparing the ground truth colors and the rendered colors:
\begin{equation}
    \mathcal{L}_\text{rgb} = \sum_{\boldsymbol{r}\in \mathcal{R}}\|\tilde{C}(\boldsymbol{r})-C(\boldsymbol{r})\|_1,
\end{equation}
where $\mathcal{R}$ denotes the set of pixels/rays in the minibatch and $\tilde{C}(\boldsymbol{r})$ is the ground-truth pixel color. 

$\mathcal{L}_\text{patch}$ is the patch warping loss following \cite{spn:6:neuralwarp,sparseneus}. According to the normal calculated at the sampled point, a local plane can be defined. By projecting the local plane to the neighboring images, the warped patch color $P_i$ for the \textit{i-th} sample is fetched. Similar to the pixel color rendering, the rendered patch color is $P(\boldsymbol{r}) = \sum_{i=1}^K T_i \alpha_i P_i$.
In this way, the patch warping loss is computed as:
\begin{align}
  \mathcal{L}_\text{patch} = \sum_{\boldsymbol{r}\in \mathcal{R}}\frac{\text{Conv}( \tilde{P}(\boldsymbol{r}), P(\boldsymbol{r}) )}
  {\sqrt{\text{Var}(\tilde{P}(\boldsymbol{r})) \text{Var}(P(\boldsymbol{r}))}},
\end{align}
where $\tilde{P}(\boldsymbol{r})$ is the ground-truth patch color. $\text{Conv}$ denotes the covariance and $\text{Var}$ denotes the variance.

$\mathcal{L}_\text{eik}$ is an eikonal term to regularize SDF values:
\begin{equation}
    \mathcal{L}_\text{eik} = \sum_{\boldsymbol{x}\in \mathcal{X} }(\| \nabla d(\boldsymbol{x}) \|_2 - 1)^2,
\end{equation}
where $\mathcal{X}$ are a set of uniformly sampled points together with near-surface points.

$\mathcal{L}_\text{normal}$ is the normal consistency loss by comparing the monocular normal prediction $\tilde{N}(\boldsymbol{r})$ and $N(\boldsymbol{r})$:
\begin{equation}
    \mathcal{L}_\text{normal} = \sum_{\boldsymbol{r} \in \mathcal{R}} \| \tilde{N}(\boldsymbol{r}) - N(\boldsymbol{r}) \|_1.
\end{equation}

\section{Experiments}

\subsection{Experimental Setting}
\begin{table*}[h!t]
\center
\footnotesize
\setlength{\tabcolsep}{4.8pt}
\begin{tabular}{c|c|cccccccccccccccc}
\toprule
Stage&\diagbox{Method}{Scan}&24&37&40&55&63&65&69&83&97&105&106&110&114&118&122&Mean \\
\midrule
\midrule

\multirow{3}*{\shortstack{3-Minutes \\ Results}}&Voxurf&1.59&0.89&0.75&0.40&1.14&0.82&0.95&1.43&1.25&0.82&0.89&1.75&0.43&0.57&0.58&0.95\\

~&NeuS2&0.79&\textbf{0.80}&1.27&\textbf{0.39}&0.95&\textbf{0.78}&0.88&1.34&1.10&0.73&\textbf{0.64}&1.19&0.45&0.58&\textbf{0.57}&0.83\\

~&Ours&\textbf{0.51}&1.11&\textbf{0.51}&0.44&\textbf{0.71}&0.98&\textbf{0.60}&\textbf{1.05}&\textbf{0.80}&\textbf{0.62}&\textbf{0.64}&\textbf{0.80}&\textbf{0.32}&\textbf{0.52}&0.60&\textbf{0.68}\\

\midrule
\multirow{8}*{\shortstack{Final \\ Reulsts}}&COLMAP&0.81&2.05&0.73&1.22&1.79&1.58&1.02&3.05&1.40&2.05&1.00&1.32&0.49&0.78&1.17&1.36 \\

~&NeRF&1.90&1.60&1.85&0.58&2.28&1.27&1.47&1.67&2.05&1.07&0.88&2.53&1.06&1.15&0.96&1.49 \\

~&SparseNeuS-F&1.71&3.38&1.75&1.00&2.64&2.42&1.07&1.54&1.56&1.21&1.23&1.05&0.74&1.24&1.24&1.58 \\

~&Instant-NGP&1.68&1.93&1.57&1.16&2.00&1.56&1.81&2.33&2.16&1.88&1.76&2.32&1.86&1.80&1.72&1.84 \\
~&VolSDF&1.14&1.26&0.81&0.49&1.25&0.70&0.72&1.29&1.18&0.70&0.66&1.08&0.42&0.61&0.55&0.86 \\

~&NeuS&0.83&0.98&0.56&0.37&1.13&\textbf{0.59}&0.60&1.45&0.95&0.78&\textbf{0.52}&1.43&0.36&\textbf{0.45}&0.45&0.77 \\

~&Voxurf&0.65&\textbf{0.74}&\textbf{0.39}&\textbf{0.35}&0.96&0.64&0.85&1.58&1.01&0.68&0.60&1.11&0.37&\textbf{0.45}&0.47&0.72\\

~&NeuS2&0.58&0.80&1.47&0.39&0.91&0.69&0.87&1.31&1.03&0.76&0.62&0.95&0.45&0.55&0.56&0.80\\

~&NeuS2*&0.56&0.76&0.49&0.37&0.92&0.71&0.76&1.22&1.08&0.63&0.59&0.89&0.40&0.48&0.55&0.70\\

~&Ours&\textbf{0.38}&1.21&0.55&0.39&\textbf{0.69}&0.93&\textbf{0.55}&\textbf{0.96}&\textbf{0.77}&\textbf{0.61}&0.55&\textbf{0.77}&\textbf{0.31}&0.49&\textbf{0.44}&\textbf{0.64}\\

\bottomrule
\end{tabular}
\caption{Quantitative evaluations on DTU dataset (lower is better). Except for the \textbf{NeuS2*} which uses different hyperparameter settings among 15 test scenes provieded by code of \cite{neus2}, all methods share the only one setting for the sake of fairness. \textbf{SparseNeuS-F} means the fine-tuning method of SparseNeuS \cite{sparseneus}. The run-time of our final result is 5 minutes.}
\label{tab:quantitative}
\end{table*}

\begin{figure*}[h!t]
\centering
\includegraphics[width=1.0\linewidth]{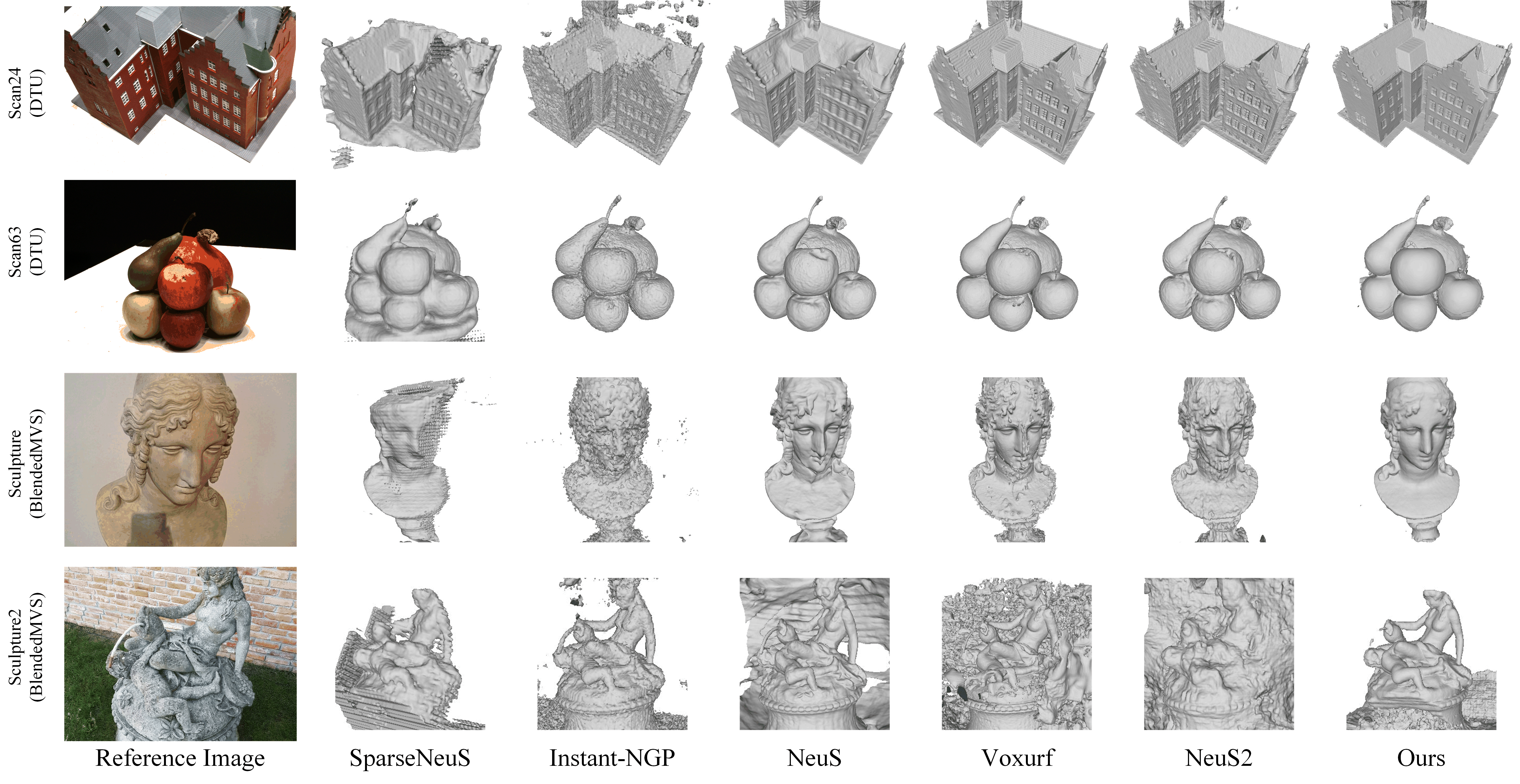}
\caption{Qualitative results on DTU and BlendedMVS dataset.}
\label{fig:quality:dtu-Blend}  
\end{figure*}

\begin{table}[htb]
\center
\footnotesize
\begin{tabular}{cc|cc|cc|cc}
\toprule
\multicolumn{2}{c|}{Voxurf} 
& 
\multicolumn{2}{c}{NeuS2} 
& 
\multicolumn{4}{|c}{Ours} \\
\midrule
CD&Time&CD&Time&CD&Time&CD&Time\\
0.72&30mins&0.80&6mins&\textbf{0.64}&5mins&0.68&\textbf{3mins}\\
\bottomrule
\end{tabular}
\caption{The comparison on surface reconstruction quality and training time among fast methods on DTU dataset.}\label{tab:time}
\end{table}

\noindent\textbf{Datasets.} We use 15 scenes of DTU dataset \cite{aanaes2016large} for quantitative and qualitative comparisons. This dataset only captures the partial geometry of objects. In addition, we also show qualitative results on several challenging $360^{\circ}$ scenes in BlendedMVS dataset \cite{yao2020blendedmvs}.

\noindent\textbf{Baselines.} We compare our method with the state-of-the-art approaches from four classes: 1) A classical traditional reconstruction method, COLMAP \cite{colmap}; 2) Per-scene optimization based neural implicit (rendering/surface reconstruction) methods with large MLPs, including NeRF \cite{nerf}, VolSDF \cite{volsdf} and NeuS \cite{neus}; 3) Per-scene optimization based neural implicit (rendering/surface reconstruction) methods with grid-based representations, including Instant-NGP \cite{ngp}, Voxurf \cite{voxurf} and NeuS2 \cite{neus2}; 4) A generic neural surface reconstruction method from sparse inputs, SparseNeuS. Note that, this method has also proposed a fine-tuning network to execute per-scene optimization. For fair comparison, in its initial stage, we use 3 adjacent images (view-id: 22, 23, 34) for all scenes rather than the carefully selected 3 images proviede by SparseNeuS. And we use all views in the scene instead of sparse views to optimize the fine-tuning network of SparseNeuS.

\begin{table}[t]
\centering
\footnotesize
\begin{tabular}{cccccc}
\toprule
&MF&RL&SS&PS&Chamfer Distance \\
\midrule
(1)&&&&&1.91\\
(2)&\checkmark& & & & 1.30\\
(3)&\checkmark&\checkmark&\checkmark&&0.84 \\
(4)&&\checkmark&&\checkmark&0.81 \\
(5)&\checkmark&\checkmark&&\checkmark&\textbf{0.64} \\
\bottomrule
\end{tabular}
\caption{\textbf{Ablation study on DTU dataset.} MF: multiple local SDF fields fusion. RL: residual learning. SS: stratified sampling. PS: prior-guided sampling.}\label{tab:ablation}
\end{table}

\begin{figure*}[t]
\centering
\includegraphics[width=\linewidth]{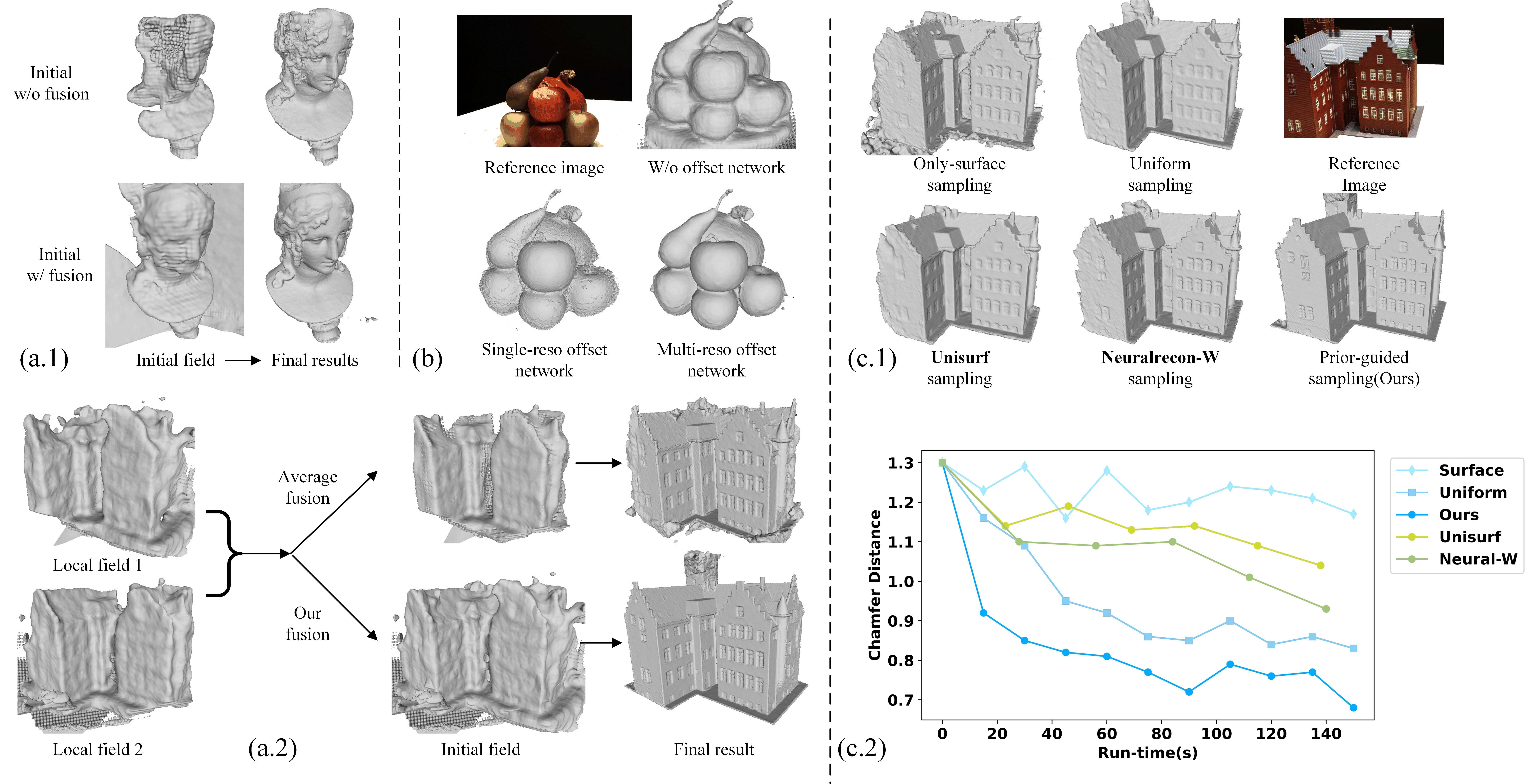}
\caption{\textbf{Ablation studies on different components.} (a.1, a.2) Multiple local SDF fields fusion. (b) SDF residual learning. (c.1, c.2) Sampling strategies. 
The ``Only surface sampling" means all sampled points locate in area $A_2$. The ``Uniform sampling" means equidistant sampling. 
The ``Unisurf sampling" means surface-based sampling in \cite{unisurf}. The ``Neuralrecon-W sampling" means hybrid voxel-and-surface-guided sampling in \cite{neuconw}. The ``Prior-guided" means our proposed prior-based sampling strategy. ``Surface'' means abbreviation of ``Only surface'', and ``Neural-W'' means that of ``Neuralrecon-W''.}
\label{fig:abl:all}
\end{figure*}

\noindent\textbf{Implementation details.} We use the pre-trained model provided by SparseNeuS without any fine-tuning to compute the initial local SDF fields. The resolution of $V_\text{global}$ is $360^3$. In Eq.~\eqref{eq:total}, $\{\lambda_1, \lambda_2, \lambda_3\} = \{1, 0.1, 0.1\}$. 
In Eq.~\eqref{eq:reserve-probability}, $\{ \beta_1,\beta_2,\beta_3\}=\{4,1,0.5\}$ 
and $N_\text{ref}=N(A_2)$. 
In the fusion stage, we use $M=2$ sparse-views groups from different perspectives to compute $V_\text{basis}$. Each sparse-views group contains 3 adjacent images.
To reconstruct a more integral initial field, the viewpoints of the initial (M=)2 sparse-view groups shall differ obviously. As 3 images in a group are adjacent , we take the viewpoint of a reference view in a group to represent the viewpoint of this group. For $360^{\circ}$ scenes on BlendedMVS dataset, we choose the two sparse-view groups whose representative viewpoints forming an angle of about $180^{\circ}$. The angle information can be computed by the provided extrinsics of reference views. For DTU dataset, the range of all viewpoints is smaller than $360^{\circ}$. We choose the two groups: group1 (view-id:22, 23, 34) and group2 (view-id:25, 26, 31) for all scenes.
In the SDF residual learning stage, to obtain the complete scene geometry, we use all views (49 on DTU dataset) to quickly optimize the rough SDF field. Moreover, if only given 3-6 sparse views rather than all views, our method can still maintain great reconstruction quality in visible area. More relative details are shown in the supplementary materials.

\noindent\textbf{Metrics.} To compare the reconstructed mesh quality among different methods, we apply the widely used chamfer distance (CD) to measure the distance between the sampled points in predicted surfaces and ground-truth points. 

\subsection{Comparisons}

\noindent\textbf{Quantitative.} The quantitative reconstruction results on DTU dataset are shown in Table~\ref{tab:quantitative}. Our method achieves the best result among all baseline methods. Meanwhile, thanks to the fused SDF field prior and the fast SDF residual learning, the results in Table~\ref{tab:time} show that our method takes only 3 minutes to reconstruct a scene, surpassing other methods.

\noindent\textbf{Qualitative.} The qualitative results on DTU and BlendedMVS dataset are presented in Fig.~\ref{fig:quality:dtu-Blend}. SparseNeuS shows over-smoothing surfaces, which lack details and contain adherent regions. The surface of NeuS and NeuS2 is clean and smooth, but in some textureless regions, the surface becomes uneven. 
To conclude, our method outperforms other methods in details and large smooth areas.

\begin{figure}[t]
\centering
\includegraphics[width=1.0\linewidth]{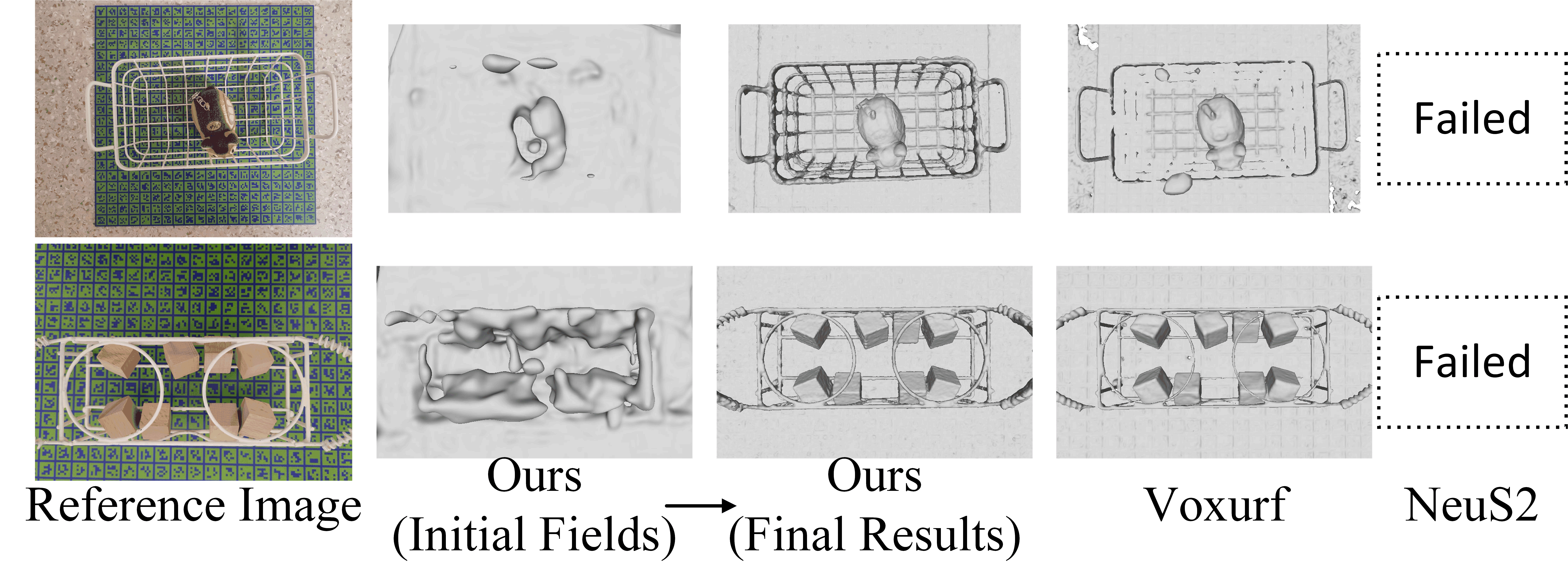}
\caption{Comprison of surface reconstruction in thin-structure scene among fast voxel-based methods.
}
\label{fig:exp:thin}
\end{figure}

\noindent\textbf{Thin-structure.} We also conduct thin-structure reconstruction comparison experiments among Voxurf \cite{voxurf} and NeuS2 \cite{neus2} in Fig.~\ref{fig:exp:thin}. For NeuS2, we use the default hyperparameter setting of DTU dataset, but fail to reconstruct meshes. We infer that good result of NeuS2 may depend on carefully modified hyperparameters, which is proved by the two different results (NeuS2 and NeuS2*) in Table \ref{tab:quantitative}.
For our method, though initial fields look quiet bad, our residual learning paradigm can still get good mesh results. Moreover, as our residual learning strategy focuses on predicting the tiny offset SDF values containing more high-frequency information, our results seem better than Voxurf.

\subsection{Ablation Studies}
We conduct ablation studies on DTU dataset to investigate the effectiveness of our proposed modules. The pretrained SparseNeuS without any fine-tuning is adopted as our baseline. Table~\ref{tab:ablation} reports the results.

\noindent\textbf{Multiple local SDF fields fusion.} We first add our fusion strategy to the baseline. As shown in Index 1 and 2 of Table~\ref{tab:ablation}, our fusion greatly improves the performance. Moreover, the baseline with our fusion has already outperformed SparseNeuS with fine-tuning (1.30 vs. 1.58). This further demonstrates the power of our fusion. In Fig.~\ref{fig:abl:all}(a.1), we apply one SDF field derived from the baseline as the initial SDF field for comparison. We see that the SDF field constructed from one sparse-views group is incomplete. Though our SDF residual learning can recover most of regions thanks to dense inputs, the completeness of the final reconstructed mesh is still unsatisfactory. Once the initial SDF field becomes more complete by fusing two complementary local SDF fields, the SDF residual learning can focus on the slight SDF offset optimization rather than the missing part recovery. Thus the final results become more complete and elaborate. This is also validated by Index 4 and 5 of Table~\ref{tab:ablation}.
Moreover, we compare our min-fusion strategy with average-fusion strategy in Fig.~\ref{fig:abl:all}(a.2). The average-fusion denotes applying the average of many local SDF values as fusion result. The initial field of average-fusion seems incomplete, as it only keeps the shared meshes among local fields, resulting in worse quality in edge areas.

\noindent\textbf{SDF residual learning.} We further adopt our SDF residual learning with stratified sampling to optimize SDF offsets. As shown in Index 2 and 3 of Table~\ref{tab:ablation}, the SDF residual learning further improves the performance. To further investigate the efficacy of SDF residual learning, we replace it by the SDF fine-tuning network used in SparseNeuS. The SDF fine-tuning network is a feature grid with resolution of $192^3$ and 8 channels. We do not apply our fusion here as the fine-tuning strategy of SparseNeuS can only work in a specific local coordinate. 
Moreover, to investigate the multi-reso's impact, we apply the single-reso feature grid with resolution of $192^3$ and 8 channels as the offset SDF network for comparison. 
As shown in Fig.~\ref{fig:abl:all}(b), the fine-tuning strategy suffers from an over-smoothing problem since it is unable to escape the local optima. The single-reso offset network is able to correct some bad surfaces, but has poor performance in smooth surfaces. The multi-reso offset network behaves well on both details and smooth surfaces.

\begin{table}[t]
\centering
\footnotesize
\begin{tabular}{cccccc}
\toprule
$\beta_1$ &:& $\beta_2$ &:& $\beta_3$ &Chamfer Distance \\
\midrule
1  &:& 1 &:& 1&0.82\\
1 &:& 4 &:& 0.5&0.95\\
0.5 &:&1  &:& 4& 0.78\\
4  &:&0.5&:& 1&0.72\\
4  &:&1  &:&0.5& \textbf{0.64}\\
\bottomrule
\end{tabular}
\caption{The comparison of different ratio settings.}
\label{tab:abl:ratio}
\end{table}

\noindent\textbf{Prior-guided sampling.} By comparing Index 3 and 5 of Table~\ref{tab:ablation}, we verify the effectiveness of our sampling strategy. Moreover, we compare it with ``only surface" sampling and ``uniform" sampling. The former only selects points locating in the area occupied by prior surfaces while the latter means equidistant sampling on a ray. 
As shown in Fig.~\ref{fig:abl:all}(c.1) and Fig.~\ref{fig:abl:all}(c.2), our sampling strategy converges quickly and achieves the best result. For ``only surface" sampling, since the prior surface has certain distance from the real surface, these sample points prevent the subsequent learning probing the real surface. Thus, it cannot converge to good results. For ``uniform" sampling, equidistant sample points in entire space lack the necessary focus on the prior-surface neighborhoods, so that its convergence speed is slower than our prior-guided sampling. 

For ``Unisurf" sampling proposed by \cite{unisurf} and ``Neuralrecon-W" sampling proposed by \cite{neuconw}, their core idea is that before applying samplings in each iteration, we shall search the surface position inferred in current field, and only distribute samples within narrow range from the inferred surface. However their samplings may not work well in fast reconstruction task, because of the expensive cost of updating current position of surface and the relatively bad quality of our initial field. Their surface-based and voxel-based samplings behaves unsatisfactory both in efficiency and final quality.

To further investigate the ratio influence of our prior-guided sampling, we adopt different ratio settings for comparison. As shown in Table~\ref{tab:abl:ratio}, if we sample most points in prior-surface areas, the results degrade rapidly from 0.64 to 0.95. The equal ratio $1:1:1$ performs the second worst as it cannot fully focus on the interesting areas provided by priors. In contrast, our default setting achieves the best.

\section{Limitation \& Future work}
The efficiency of per-iteration optimizaiton of our residual learning is limited, which is 2x slower than Voxurf \cite{voxurf} and 3x slower than NeuS2 \cite{neus2}, as our current method is based on the Pytorch rather than Cuda and C++. We plan to rebuild our method based on the great project NeuS2 \cite{neus2}, so as to achieve faster reconstruction within 2 minutes.

\section{Conclusion}
We have proposed a prior-based residual learning paradigm for fast multi-view neural surface reconstruction. Our proposed multiple local SDF fields fusion can quickly obtain a more integral global SDF field priors from different local field priors. Moreover, to accelerate the optimization, we present the prior-guided sampling strategy to distribute more sampling points in interesting areas. Besides, by designing a lightweight offset network to predict SDF offsets, our method achieves fast and high-quality surface reconstruction within 3 minutes. Extensive experiments show the superior performance of PR-NeuS on various datasets.

{
    \small
    \bibliographystyle{ieeenat_fullname}
    \bibliography{main}
}


\end{document}